\documentclass[11pt]{article}

\usepackage[preprint]{acl}

\usepackage{times}
\usepackage{amssymb}
\usepackage{amsmath}
\usepackage{latexsym}
\usepackage{multirow}
\usepackage{booktabs}
\usepackage{colortbl}
\usepackage{tabularx}
\usepackage{adjustbox}
\usepackage{subcaption}
\usepackage{algpseudocode}
\usepackage{hyperref}
\usepackage{dsfont}
\usepackage{algpseudocode}
\usepackage{arydshln}
\usepackage{tikz}
\usetikzlibrary{patterns.meta}
\usepackage{hhline}

\usepackage{tcolorbox}
\tcbuselibrary{skins,breakable,theorems}
\usepackage[T1]{fontenc}

\usepackage[utf8]{inputenc}

\usepackage{microtype}

\usepackage{inconsolata}

\usepackage{graphicx}

\definecolor{dkblue}{rgb}{0,0,0.5}
\newcommand{\splus}[1]{\scalebox{0.75}{\textcolor{red}{(+#1)}}}%
\newcommand{\sneg}[1]{\scalebox{0.75}{\textcolor{red}{(-#1)}}}%
\newcommand{\qwenone}{Qwen-2.5 1.5B\space}
\newcommand{\qwenseven}{Qwen-2.5 7B\space}
\newcommand{\qwenfrtn}{Qwen-2.5 14B\space}

\newcommand{\smallsym}[2]{#1{\mathpalette\make@small@sym{#2}}}

\newtcolorbox[auto counter]{example}[2][]{
  width=\textwidth,
  sharp corners=all,
  colback=white!96!black,
  float*,
  floatplacement=t,
  label=#2,
  title={Box \thetcbcounter. #1},
}

\title{Cross-lingual Self-Consistency for Multilingual Reasoning\\with Language Models}

\author{Ahmed Elhady$^{1}$ \quad Eneko Agirre$^{1}$ \quad Mikel Artetxe$^{1,2}$ \\
$^{1}$HiTZ Center, University of the Basque Country (UPV/EHU) \qquad $^{2}$Reka AI \\
\texttt{\{ahmed.salemmohamed,e.agirre,mikel.artetxe\}@ehu.eus} }

\begin{document}
\maketitle
\begin{abstract}

Despite expanding their multilingual coverage, the advanced reasoning capabilities of LLMs remain largely confined to a few high-resource languages like English. To address this, we propose an \textit{unsupervised} Reinforcement Learning (RL) approach to enhance multilingual reasoning by enforcing cross-lingual self-consistency---the principle that a model should produce the same final answer for equivalent problems in different languages. Existing methods are limited by the scarcity of multilingual reasoning data, and show weak generalization to unseen languages. Our approach requires neither gold answers nor parallel data, and it achieves average gains of up to \textbf{21.7\%} on MGSM across 10 languages. In addition, our method demonstrates strong generalization, with an \textbf{18.2\%} mean improvement on MGSM languages unseen during training, and up to \textbf{6.2\%} gain on 3 out-of-distribution benchmarks. These results show the potential of consistency-based methods to improve the multilingual capabilities of LLMs without requiring supervised data.
\end{abstract}

\section{Introduction}

Recent advances in Large Language Models (LLMs) show a promising trend to support a wider range of languages. For example, the pretraining of Qwen3 models \cite{yang2025qwen3technicalreport} included over 100 languages compared to 29 for Qwen2.5 \cite{qwen2.5}. Nevertheless, their performance is limited to basic capabilities, while their effectiveness in tasks that require complex reasoning remains concentrated in a limited set of languages, such as English and Chinese \cite{ghosh2025multilingualmindsurvey}. 

Consequently, prior work predominantly leverages the reasoning on a strong language, typically English. For instance, given a question in the source language, the model can be prompted to reason in an English chain-of-thought instead of the native language \cite{liu-etal-2025-translation, zhu2024question}. As an alternative,  the source-language query is translated into English in a first step, to then perform the reasoning in English in a second step. Translations can be provided by the model itself (self-translation), or, alternatively, by an external machine translation engine \cite{liu-etal-2025-translation}. Though simple, these strategies limit applicability in realistic scenarios where reasoning steps must be expressed in the source language.

\begin{figure}[t!]
\includegraphics[width=1.\columnwidth]{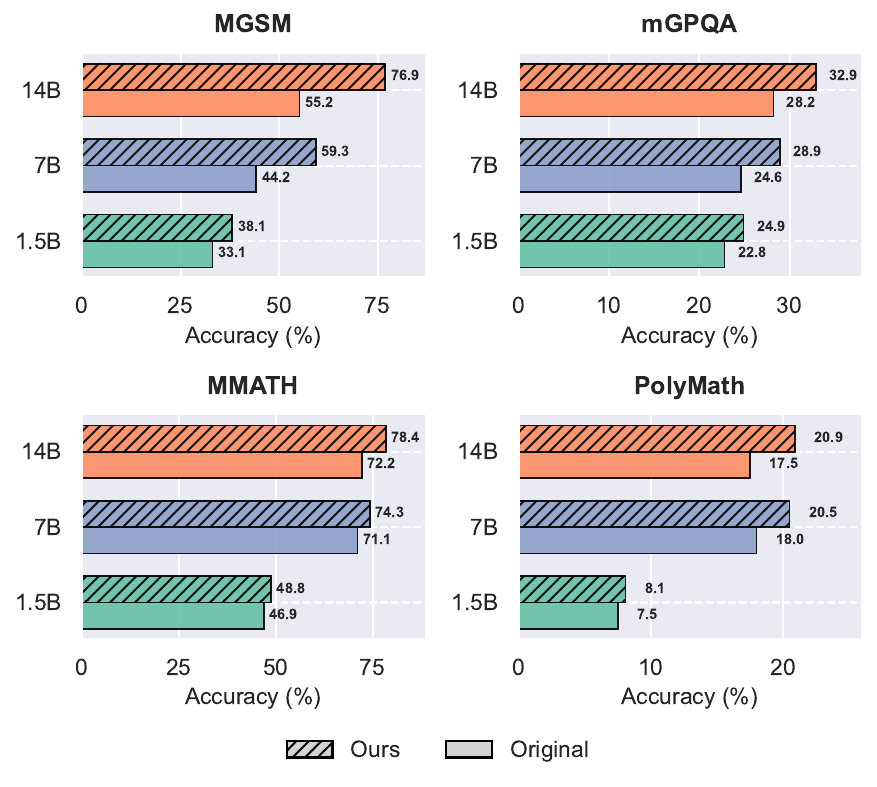}
\caption{\textbf{Accuracy of Qwen2.5 Instruct.} Our method significantly improves over the original model, while only requiring English questions for training (without gold answers or pre-existing translations).}
\label{fig:main_results}
\end{figure} 

\begin{figure*}[th!]
\centering
\includegraphics[width=1.0\linewidth]{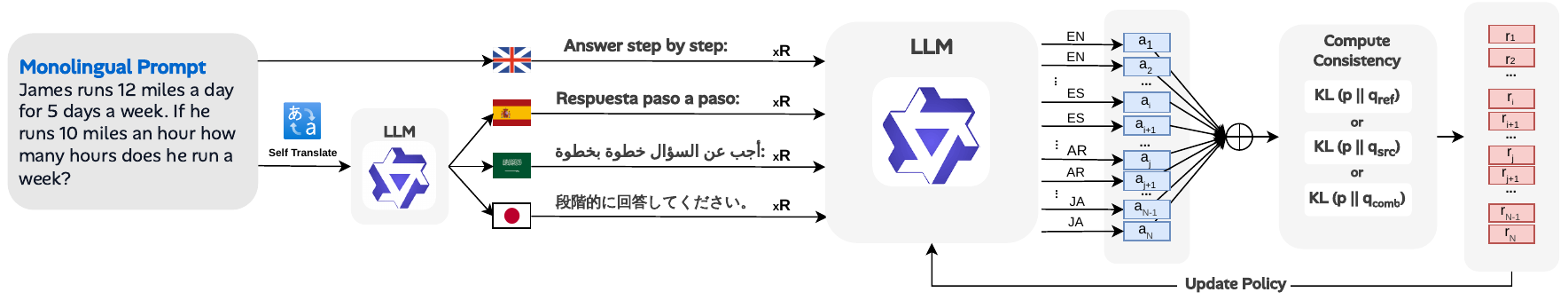}
\caption{\textbf{Overview of our cross-lingual self-consistency framework}. Each monolingual prompt is self-translated by the model into the target languages and answered in every language, yielding set of outputs Y. Rewards are computed as the consistency between the output distribution $p(y)$ and a target distribution $q(y)$ — estimated from the reference language ($q_{ref}$), the prompt language ($q_{src}$), or all outputs combined ($q_{comb}$).}
\label{fig:main-app}
\end{figure*}

In order to reduce the performance gap when reasoning in the source language, prior work focused on improving the cross-lingual alignment. 
Some works use supervised fine-tuning (SFT) on translations of questions to and from the reference languages \cite{zhu2024question} 
or control the response generation by manipulating the logits to mimic those of English \cite{wang-etal-2025-bridging, huang2024mindmergerefficientboostingllm}.
Other works use post-training methods, namely DPO \cite{rafailov2023direct}, to push models to generate correct reasoning traces translated from English and discourage incorrect, natively generated ones \cite{yang2025language, lee2025crosslingualoptimizationlanguagetransfer}. 
These approaches rely on translated questions and reasoning traces for supervision and, as a result, generalization to unseen languages and domains is constrained by both machine-translation quality and the scarcity of high-quality manually translated data
\cite{ghosh2025multilingualmindsurvey}.

In this paper, we propose to use cross-lingual self-consistency as an \textbf{unsupervised} signal for RL. 
Our method extends existing monolingual self-consistency approaches, which capture the intuition that different reasoning paths should lead to consistent final answers, to further push the model to produce the same final answer for equivalent problems in different languages (Figure \ref{fig:main-app}). This design guarantees broad practical applicability, since it requires \textbf{no access to ground truth labels nor pre-translated reasoning traces}. This is specially useful given that multilingual reasoning data is only available for evaluation, and training resources are largely monolingual \cite{ghosh2025multilingualmindsurvey}. Our experimental setup only uses English questions. 

We demonstrate the effectiveness of our method across three model scales, achieving an \textbf{average performance improvement of 7.8\% to 21.7\%} over 10 languages on the MGSM benchmark (Figure \ref{fig:main_results}). The improvements can be amplified when high-quality translations for the questions are available (\S\ref{sec:translation-quality}). Furthermore, our approach exhibits strong generalization, yielding up to \textbf{18.2\% average improvement on unseen languages} within MGSM and extending effectively to other domains with a 4.7\% average gain on the mGPQA \cite{huang2025benchmax} and 6.8\% on MMATH \citep{luo2025mmathmultilingualbenchmarkmathematical} benchmarks (\S\ref{sec:gen}). Collectively, these results show that our method is effective at narrowing the multilingual reasoning gap in LLMs, advancing their robustness and equity across languages. 

\section{Background} \label{sec:self-cons-bkgrnd}

Recent work has shown improvements in the monolingual reasoning capabilities of LLMs by maximizing the consistency of the model \cite{he2026farunsupervisedrlvrscale,prabhudesai2025maximizing, chen2025selfquestioninglanguagemodels, agarwal2025unreasonableeffectivenessentropyminimization, cui2025entropymechanismreinforcementlearning}. We next formalize this framework, and present our proposed extension to the multilingual scenario in \S\ref{sec:reward-modeling}.

Given an input $x$, let $\{(c_i, y_i) \sim \pi(\cdot \mid x) \}_{i=1}^N$ denote a set of $N$ trajectories sampled from the policy model $\pi$, where $c_i$ is the chain-of-thought reasoning and $y_i$ is the final answer following it. The core idea behind these methods is that, even if the reasoning paths $c_i$ can differ, a strong model should always arrive at the same (ideally correct) final answer $y_i$. This principle of final answer consistency can be optimized through RL, even if the ground truth answer is not known. To that end, existing methods minimize the entropy of the marginalized answer distribution.

More precisely, let $\hat{p}_{\pi}(y|x)$ be the empirical probability of a specific final answer $y$:
\begin{equation}
    \hat{p}_{\pi}(y|x) = \frac{1}{N}\sum_{i=1}^N \mathbf{1} (y_i=y)
\end{equation}
The aggregated reward over all sampled generations of $x$ is the negative of the entropy of this distribution:
\begin{equation} \label{eq:r_total_mono}
\begin{split}
    R_{\mathrm{total}}(x) &= -H(\hat{p}_\pi (\cdot|x)) \\
    &= \sum_{y} \hat{p}_\pi(y|x) \log \hat{p}_\pi(y|x)
\end{split}
\end{equation}
and the reward of each individual generation is decomposed as follows:
\begin{equation}
    R(x, c_i, y_i) = \log \hat{p}_\pi(y_i|x)
\end{equation}
so it holds that $R_{\mathrm{total}}(x)=\frac{1}{N}\sum_{i=1}^N R(x, c_i, y_i)$. This reward function is optimized through RL, with GRPO \cite{shao2024deepseekmathpushinglimitsmathematical} being a common choice.

\section{Proposed Method}
\label{sec:reward-modeling}

In the multilingual scenario, not only do we want models to produce consistent final answers within each language, but we also want them to produce consistent final answers for equivalent inputs in different languages. We next propose a multilingual extension of the framework in \S\ref{sec:self-cons-bkgrnd} that formalizes this principle of cross-lingual self-consistency.

Let $\mathbf{x}=\{x^{(l)}\}_{l=1}^L$ be a set of parallel inputs in $L$ languages, and let $\{(c_i^{(l)}, y_i^{(l)}) \sim \pi(\cdot \mid x^{(l)}) \}_{i=1}^N$ be a set of $N$ generations for the input in language $l$. We assume the final answers $y_i^{(l)}$ are normalized into a shared canonical form (e.g., numeric values), so that equality is well-defined across languages. We define the following aggregated reward for the input in each language $x^{(l)}$, which replaces the entropy term in Equation \ref{eq:r_total_mono} with the cross-entropy with respect to a second distribution $\hat{q}_\pi$:
\begin{equation}
\begin{split}
    R_{\mathrm{total}}(x^{(l)}) &= -H \left(\hat{p}_\pi (\cdot|\mathbf{x}, l), \hat{q}_\pi (\cdot|\mathbf{x}, l) \right) \\
    &= \sum_{y} \hat{p}_\pi(y|\mathbf{x}, l) \log \hat{q}_\pi(y|\mathbf{x}, l)
\end{split}
\end{equation}
where $\hat{p}_{\pi}(y|\mathbf{x}, l)$ is the empirical marginalized answer distribution in language $l$:
\begin{equation}
    \hat{p}_{\pi}(y|\mathbf{x}, l) = \frac{1}{N}\sum_{i=1}^N \mathbf{1} \left( y_i^{(l)}=y \right)
\end{equation}
and $\hat{q}_{\pi}(y|\mathbf{x}, l)$ is the empirical target distribution, for which we define several alternatives below. Replacing entropy with cross-entropy allows each language-specific answer distribution to be aligned to a target distribution derived from other languages, thereby operationalizing cross-lingual consistency.
The reward for each sampled generation is given by
\begin{equation}
    R \left(x^{(l)}, c_i^{(l)}, y_i^{(l)} \right) = \log \hat{q}_\pi \left(y_i^{(l)}|\mathbf{x}, l\right)
\end{equation}
so, analogous to the monolingual case, it holds that 
$R_{\mathrm{total}}(x^{(l)})=\frac{1}{N}\sum_{i=1}^N R(x^{(l)}, c_i^{(l)}, y_i^{(l)})$.

In our experiments, we explore the following alternatives for $\hat{q}_\pi$, which differ on the language(s) they use to estimate the target distribution:
\begin{enumerate}
    \item \textbf{The source language $l$:} In this case, the source and target distributions are the same:
    \begin{equation}
        \hat{q}_{\pi}(y|\mathbf{x}, l) = \hat{p}_{\pi}(y|\mathbf{x}, l)
    \end{equation}
    Given that $H(p, p)=H(p)$, this is equivalent to minimizing the entropy of the marginalized answer distribution for each language without any constraint on cross-lingual consistency. When $L=1$, this further becomes equivalent to the formulation in \S\ref{sec:self-cons-bkgrnd}, making our framework a strict generalization of monolingual self-consistency.

    \item \textbf{The reference language $\mathrm{en}$:} We take the marginalized answer distribution in another (reference) language, which, without loss of generality, we set to English:\footnote{In practice, one would choose the language that is expected to have the strongest monolingual performance which, in most scenarios (including ours), is English.}
    \begin{equation}
        \hat{q}_{\pi}(y|\mathbf{x}, l) = \max \left(\hat{p}_{\pi}(y|\mathbf{x}, \mathrm{en}), \epsilon\right)
    \end{equation}
    $\epsilon$ prevents the reward becoming $-\infty$ when $\hat{p}_{\pi}(y|\mathbf{x}, \mathrm{en})=0$ and $\hat{p}_{\pi}(y|\mathbf{x}, l)>0$, while still strongly favoring answers that are probable in English.\footnote{We set $\epsilon=0.01$ in our experiments, but find that it is rarely triggered in practice.} This variant can be seen as a form of cross-lingual distillation that, by aligning all languages to the same reference, indirectly makes them more consistent with one another.

    \item \textbf{All languages combined:} We take the empirical answer distribution marginalizing over all languages and trajectories:
    \begin{equation}
        \hat{q}_{\pi}(y|\mathbf{x}, l) = \frac{1}{L} \sum_{k=1}^L\hat{p}_{\pi}(y|\mathbf{x}, k)
    \end{equation}
    This variant directly optimizes cross-lingual self-consistency by pushing the final answer distribution in each language to match the aggregated one across all languages. Unless otherwise indicated, we use this variant in our experiments.
\end{enumerate}

\begin{table*}[t!]
\small
\centering

\begin{subtable}{\linewidth}
\centering
\begin{adjustbox}{max width=1.0\linewidth}
\begin{tabular}{lrrrrrrrrrrr!{\vrule}l}
\bottomrule
\rowcolor{gray!15} 
& \textbf{bn} & \textbf{de} & \textbf{en} & \textbf{es} & \textbf{fr} & \textbf{ja} & \textbf{ru} & \textbf{sw} & \textbf{te} & \textbf{th} & \textbf{zh} & \textbf{avg}\\
\toprule
\textbf{\qwenone IT}&12.0&41.2&52.4&48.8&48.1&31.6&46.5&2.8&4.4&34.6&41.2&33.1\\
\quad trans-test &14.4&44.2&52.4&50.0&42.1&23.6&38.8&8.4&\textbf{12.4}&36.8&40.8&33.4 \splus{0.3}\\
\quad LIDR \citep{yang2025language}&18.0&52.0&\textbf{57.6}&54.4&54.0&\textbf{39.6}&\textbf{55.6}&\textbf{8.4}&6.4&37.2&\textbf{55.6}& 39.9 \splus{6.8}\\
\quad S1 \citep{yong2025crosslingualreasoningtesttimescaling}& 17.2&47.6&54.1&57.2&52.8&32.8&50.0&6.0&4.4&40.7&50.4&37.6 \splus{4.5}\\
\quad self-cons (ours)&\textbf{21.2}&\textbf{53.6}&50.3&\textbf{63.2}&\textbf{56.4}&36.0&\textbf{55.2}&5.2&7.2&\textbf{48.0}&54.0&\textbf{40.9} \splus{7.8}\\
\midrule
\textbf{\qwenseven IT}&30.0&60.4&78.6&40.2&48.2&49.8&37.4&19.8&18.2&54.8&49.2&44.2\\
\quad trans-test&33.4&67.6&78.6&50.3&54.4&43.6&39.2&11.4&18.6&60.2&52.6& 46.4 \splus{2.2}\\
\quad LIDR \citep{yang2025language}&51.2&68.8&79.6&60.4&66.4&60.4&55.2&\textbf{25.2}&17.2&56.0&60.0&54.6 \splus{10.3}\\
\quad S1 \citep{yong2025crosslingualreasoningtesttimescaling}& 50.0&63.6&\textbf{88.2}&66.4&70.0&\textbf{63.2}&72.4&8.8&15.6&64.4&\textbf{60.4}&55.9 \splus{11.7} \\
\quad self-cons (ours)
&\textbf{54.8}&\textbf{75.2}&81.8&\textbf{72.4}&\textbf{71.6}&55.6&\textbf{76.0}&17.6&\textbf{24.0}&\textbf{73.2}&59.6& \textbf{59.3} \splus{15.1}\\
\midrule
\textbf{\qwenfrtn IT}&37.6&61.6&84.2&75.2&48.8&47.2&44.0&40.4&21.8&66.0&80.8&55.2\\
\quad trans-test&33.4&66.0&84.2&70.4&57.3&53.2&49.2&41.4&26.2&70.2&82.3& 57.6 \splus{2.4}\\
\quad LIDR \citep{yang2025language}&54.0&71.2&87.6&81.2&65.2&51.6&66.8&50.8&38.6&69.8&86.0& 65.7 \splus{10.5}\\
\quad S1 \citep{yong2025crosslingualreasoningtesttimescaling}&60.4 & 86.4 & \textbf{\underline{90.4}}&86.8& 80.4&\textbf{\underline{80.8}}&\textbf{\underline{89.2}}&\textbf{\underline{54.0}}&40.4&76.0&85.2&  75.6 \splus{20.4} \\
\quad self-cons (ours)&\textbf{\underline{65.6}}&\textbf{\underline{89.2}}&88.4&\textbf{\underline{87.8}}&\textbf{\underline{82.2}}&\textbf{\underline{80.8}}&88.4&49.6&\textbf{\underline{44.6}}&\textbf{\underline{81.6}}&\textbf{\underline{88.0}}& \textbf{76.9} \splus{21.7}\\
\bottomrule
\end{tabular}
\end{adjustbox}
\caption{\textbf{Qwen2.5 baselines}}.
\label{tab:main-res-qwen}
\end{subtable}

\begin{subtable}{\linewidth}
\centering
\begin{adjustbox}{max width=1.0\linewidth}
\begin{tabular}{lrrrrrrrrrr!{\vrule}l}
\bottomrule
\rowcolor{gray!15} 
& \textbf{bn} & \textbf{de} & \textbf{en} & \textbf{es} & \textbf{fr} & \textbf{ja} & \textbf{ru} & \textbf{sw} & \textbf{th} & \textbf{zh} & \textbf{avg}\\
\toprule
\textbf{Llama3 8B-IT}&42.4&58.8&73.2&64.4&57.2&40.8&58.8&31.6&35.6&49.6&51.2\\
\quad MAPO \citep{she-etal-2024-mapo}\textsuperscript{\textdagger}& 38.4 & 50.8 & 71.6 & 62.8 & 59.2 & 38.4 & 62.0 & 30.8 & 47.2 & 54.0 & 51.5 \splus{0.3}\\
\quad LIDR \citep{yang2025language}& 47.2 & 57.0 & 72.0 & 64.0 & 60.8 & 47.6 & 62.0 & 32.8 & \textbf{\underline{53.6}} & 59.2 & 55.6 \splus{4.4}\\
\quad self-cons (ours)&\textbf{\underline{64.4}}&\textbf{\underline{76.4}}&\textbf{\underline{82.4}}&\textbf{\underline{74.8}}&\textbf{\underline{72.0}}&\textbf{\underline{57.2}}&\textbf{\underline{69.6}}&\textbf{\underline{59.2}}&41.6&\textbf{\underline{68.4}}&\textbf{65.5} \splus{14.3}\\
\midrule
\textbf{Llama2-13B}&35.2&44.4&53.2&48.0&48.4&43.2&47.6&42.8&46.8&48.8&45.8\\
\quad MathOctopus-13B \citep{chen-etal-2024-breaking}\textsuperscript{\textdagger}&42.0&49.2&51.6&53.2&49.6&39.6&47.6&46.0&46.0&51.2&47.6 \splus{1.8}\\
\quad MetaMathOctopus-13B \citep{she-etal-2024-mapo}\textsuperscript{\textdagger}&41.6&\textbf{60.1}&66.8&61.1&60.8&57.3& 59.1&50.9&52.1&53.1&51.5 \splus{5.7}\\
\quad MAPO-13B \citep{she-etal-2024-mapo}\textsuperscript{\textdagger}&\textbf{44.8}&59.2&\textbf{71.6}&\textbf{62.8}&\textbf{63.6}&\textbf{56.0}&\textbf{59.3}&\textbf{55.2}&\textbf{47.6}&\textbf{59.6}&\textbf{58.0} \splus{12.2}\\
\midrule
\textbf{Llama2-7B}&18.8&38.0&52.0&39.2&36.4&27.2&33.6&23.6&21.6&31.6&32.2\\
\quad INCLINE-7B \citep{wang-etal-2025-bridging}\textsuperscript{\textdagger}&\textbf{48.4}&\textbf{51.2}&46.8&\textbf{45.2}&37.6&\textbf{44.8}&38.0&\textbf{38.8}&\textbf{43.2}&42.8&\textbf{43.7} \splus{11.5}\\
\quad MathOctopus-7B \citep{chen-etal-2024-breaking}\textsuperscript{\textdagger}&33.2&43.6&\textbf{54.8}&\textbf{45.2}&\textbf{38.0}&35.6&\textbf{48.4}&38.4&36.4&\textbf{45.2}&41.9 \splus{9.7}\\
\bottomrule
\end{tabular}
\end{adjustbox}
\caption{\textbf{Prior published methods.}}
\label{tab:main-res-prior}
\end{subtable}

\caption{\textbf{Main results on MGSM (accuracy).} Our method outperforms all Qwen2.5 baselines (a), with gains growing as model size increases, and exceeds prior published methods (b). All Qwen baselines are trained by us. We underline the overall best performance for each language.\textsuperscript{\textdagger} indicates results reported by the method authors}
\label{tab:main-res}
\end{table*}

Our approach is \textbf{unsupervised} in that it does not require any parallel data labeled by humans. To obtain these, we adopt \textbf{self-translation}: starting from a set of English questions, we prompt the model itself to translate them into other languages. Although self-translation is noisy, we adopt it for its practicality and scalability, while avoiding external translation systems that could inject additional knowledge into the model. In practice, we find that aggregating multiple generations—e.g., via majority voting at high sample counts—substantially mitigates translation errors and yields reliable baseline performance (see App.~\ref{app:maj@k} for details).\footnote{Training on noisy signals is common in machine learning and can improve generalization. For example, although applying dropout at test time can hurt performance, it remains a standard and effective training technique.}

\section{Experimental Setup}

\paragraph{Base Model} We conduct our experiments and ablations on the Qwen-2.5 family of models \cite{qwen2.5}. We chose those models since they (i) are pre-trained and supervised fine-tuned on over 29 languages, (ii) come in a handful of practical sizes, facilitating ablations at scales, and (iii) widely used in multilingual research. In addition, we experiment with the Llama3 8B model \cite{grattafiori2024llama3herdmodels} to compare with prior work that use Llama-based models. 

\paragraph{Baselines}
We evaluate our approach against two categories of baselines.

\textbf{(A) Prompt-based methods.} We consider two prompting strategies: Original model performance using \space\emph{native-cot}, where the model performs reasoning and produces answers in the original language of the query, and \emph{translate-test}, where the query is first translated into English and the model subsequently solves the problem in English \cite{gao2025thinkingmultilinguallyempowerllm}. 

\textbf{(B) Finetuning methods.} We include two of the best-performing approaches that improve reasoning via finetuning. The first is test-time scaling (S1) \cite{yong2025crosslingualreasoningtesttimescaling}, which applies SFT on English data to enhance reasoning capabilities. The second is LIDR \cite{yang2025language}, which leverages DPO to improve multilingual reasoning performance. We train both methods on the Qwen2.5 models ourselves using the official code bases. In addition, we report relative performance gains compared to the prior published methods that use Llama base models.

\paragraph{Monolingual Training Dataset} Our method only requires English questions for training. We randomly sample 1k questions from the training split of the English-only GSM8k dataset \cite{cobbe2021gsm8k}. 
The final output is parsed using the standard regular expressions from the original benchmark.
As discussed in \S\ref{sec:reward-modeling}, we use self-translation to obtain parallel questions, and analyze the impact of manual translation in \S\ref{sec:translation-quality}

\paragraph{Evaluation Data} Main evaluation and ablations are done on MGSM \cite{shi2022languagemodelsmultilingualchainofthought}, which consists of 250 math word problems in 10 languages, which were sampled and manually translated from the GSM8K test set. In addition, we evaluate the generalization capabilities on 3 OOD datasets: mGPQA \cite{huang2025benchmax}, MMath \cite{luo2025mmathmultilingualbenchmarkmathematical}, and PolyMath \cite{wang2025polymath}.

\paragraph{Training Environment and Hyperparameters} We use GRPO as the RL algorithm and use VeRL \cite{sheng2024hybridflow} for the training backend. We use a batch size of 32 parallel prompt sets (making a total of 320 prompts per step across all the 10 languages), with 32 rollouts per language and prompt, resulting in a total of 160 steps. The mini-batch size is set to 64 prompts (5 mini-batches per step), the KL coefficient to 0.001 and the learning rate to 1E-06. We set the temperature to 0.4.\footnote{Setting the temperature to a higher value sometimes led to unstable training. We leave investigating this for future work.} All models are trained using 4$\times$A100 GPUs. Other hyperparameters are left at the default values.

\section{Results}

\paragraph{Main Results}

Table~\ref{tab:main-res} contains the main results of our method. As show in Table~\ref{tab:main-res-qwen}, we achieve average gains of 7.8, 15.1, and 21.7 points in Qwen2.5-IT 1.5B, 7B, and 14B, respectively, with improvements across nearly all languages. Notably, gains increase consistently with model size. We note that in \qwenfrtn, Russain (ru) and Telugu (te) achieve the highest relevant performance gains: 102\% and 104\% respectively. We also observe degradation in Swahili for \qwenseven and in English and Telugu in the \qwenone model. 

Models trained with S1 test-time scaling remain competitive with our approach. In particular, they outperform our method on 3 languages for both \qwenfrtn and \qwenseven. However, these gains come at the cost of substantially longer reasoning traces, requiring approximately 3× more tokens. In contrast, our method achieves stronger overall performance with considerably more efficient token usage (see App.~\ref{app:token-budget}). Furthermore, the improvements from S1 are driven primarily by the \textit{quote-and-think} paradigm \citep{yong2025crosslingualreasoningtesttimescaling}, where the model performs intermediate reasoning in English before generating the final answer in the target language. Consequently, these gains do not necessarily indicate improved intrinsic multilingual reasoning ability (see App.~\ref{app:quote-and-think}).

\begin{figure}[t!]
\includegraphics[width=1.0\columnwidth]{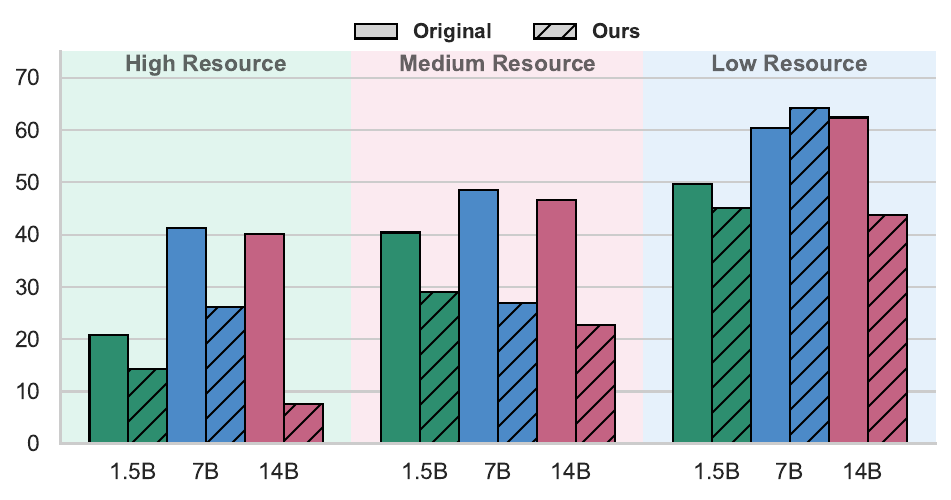}
\caption{\textbf{Maximum performance difference relative to English within each resource tier on MGSM for Qwen2.5 models.}.  Our method significantly reduces the performance gap, with exception in the low-resource languages of the 7B model.}
\label{fig:perf-variance}
\end{figure}

\paragraph{Performance Difference w.r.t English}
English is the best-performing language in the all baseline models. We examine the performance difference between English and other languages before and after training with our method. We partition languages into high-, medium-, and low-resource groups according to their representation in Common Crawl archives.\footnote{\url{commoncrawl.github.io/cc-crawl-statistics/plots/languages}} and show the maximum performance gap within each tier in Figure~\ref{fig:perf-variance}. Notably, the gap in performance is reduced up to 4 folds in High-resource languages for the \qwenfrtn model. This suggests that cross-lingual performance gaps are not solely attributable to differences in factual or linguistic coverage, but may instead arise from inconsistencies in multilingual reasoning behavior. We observe similar behavior in the Medium-resource tier. Across model sizes, our method yields a comparable performance gap of 23-28\% in the Medium-resource group. We observe that the difference increases in Low-resource languages for the \qwenseven model, which can be explained by the degradation observed in Swahili for that particular model.

Although our method yields the largest gains for low-performing languages, they still lag behind high-resource languages. This reflects limitations of the multilingual backbone model (Qwen), whose weaker language proficiency in these languages constrains our method's efficiency.

\paragraph{Comparison to Prior Methods}
\label{sec:comparison}
Head-to-head comparison with alternative methods  to improve multilingual reasoning is obscured by different test sets, languages or base models. In Table~\ref{tab:main-res-prior} we report results on MGSM or other methods including INCLINE \cite{wang-etal-2025-bridging}, Mathoctopus \cite{chen-etal-2024-breaking}, MetaMathOctopus \cite{she-etal-2024-mapo}, LIDR \cite{yang2025language} and MAPO \cite{she-etal-2024-mapo}.\footnote{Note that Telugu was not reported for any prior model.} The methods are applied to Llama2 or Llama3 base models. We run our self-consistency method on Llama3-8B model, and obtain the best results on average. 

\section{Analysis and Discussion}

Below we analyze whether our method improves the cross-lingual self-consistency of the model, whether it generalizes to unseen domains and languages, the impact of the target distribution and that of multilingual datasets. 
\begin{figure}[t!]
\centering
\includegraphics[width=1.\columnwidth]{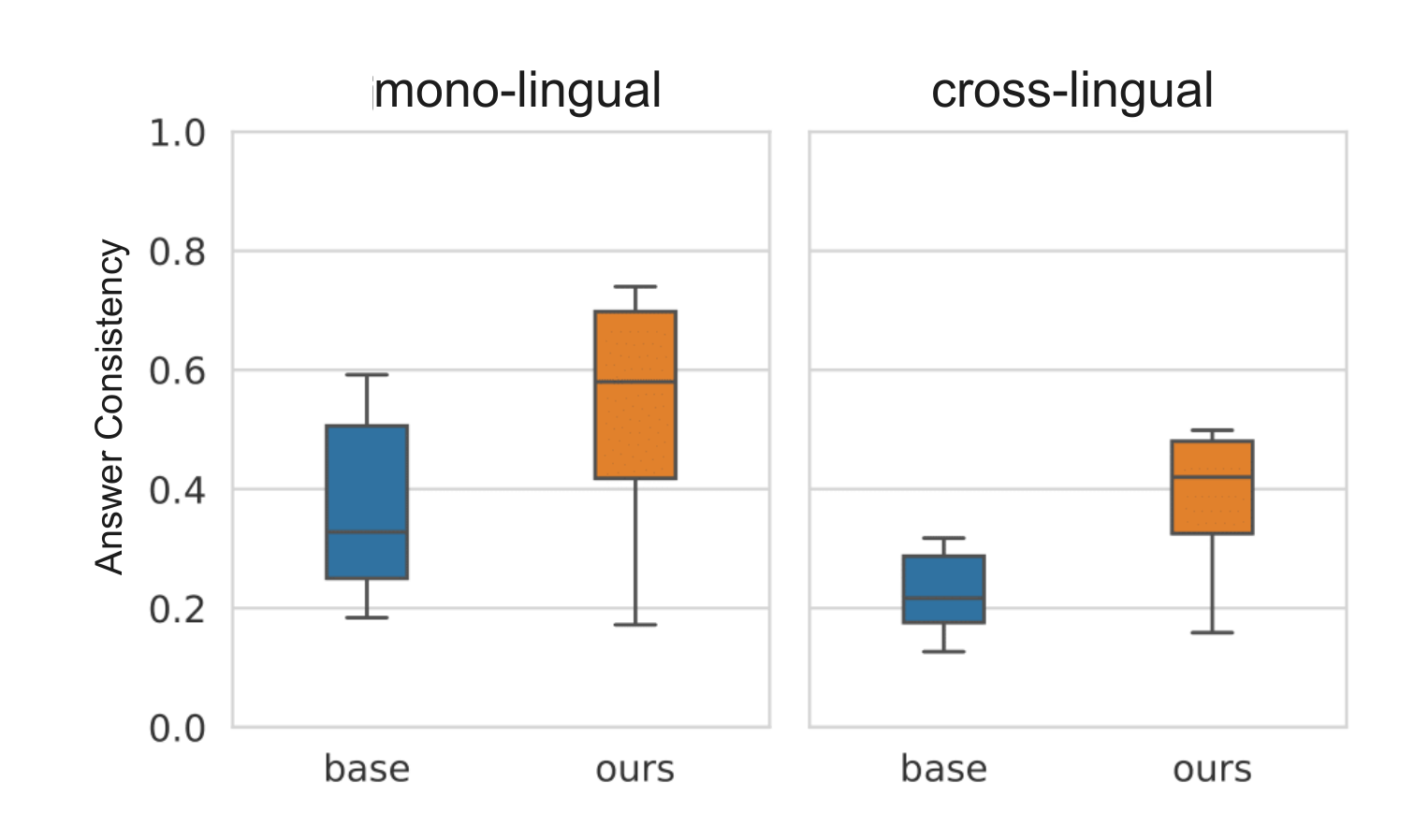}
\caption{\textbf{Monolingual and Cross-lingual consistency} of \qwenseven model on the MGSM benchmark. Our method improves both monolingual and cross-lingual consistency of the model.}
\label{fig:qwen7b-cons}
\end{figure}

\subsection{Cross-Lingual Self-Consistency}

Monolingual self-consistency methods work by minimizing the entropy of the marginalized answer distribution (see \S\ref{sec:self-cons-bkgrnd}). For easier interpretability, we next analyze the monolingual self-consistency of different models by taking the frequency of the most repeated answer instead:
\begin{equation}
    \max_i \frac{1}{N} \sum_{j=1}^N \mathbf{1} \left( y_i^{(l)}=y_j^{(l)} \right)
\end{equation}
Similarly, we measure cross-lingual self-consistency for each language $l$ by taking the maximum frequency with which any of its answers appears across the rest of the languages:
\begin{equation}
    \max_i \frac{1}{N (L-1)} \sum_{\substack{k=1 \\ k \ne l}}^L \sum_{j=1}^N \mathbf{1} \left( y_i^{(l)}=y_j^{(k)} \right)
\end{equation}
We analyze self-consistency on MGSM before and after applying our method to \qwenseven. To that end, we sample 32 parallel responses per example and language, and compute the average monolingual and cross-lingual self-consistency for each language. As shown in Figure~\ref{fig:qwen7b-cons}, our approach greatly improves both consistency measures, with monolingual consistency remaining higher than cross-lingual consistency. In Appendix~\ref{app:training-curves} we show that self-consistency improves as training progresses.

\begin{figure}[t!]
\includegraphics[width=1.\columnwidth]{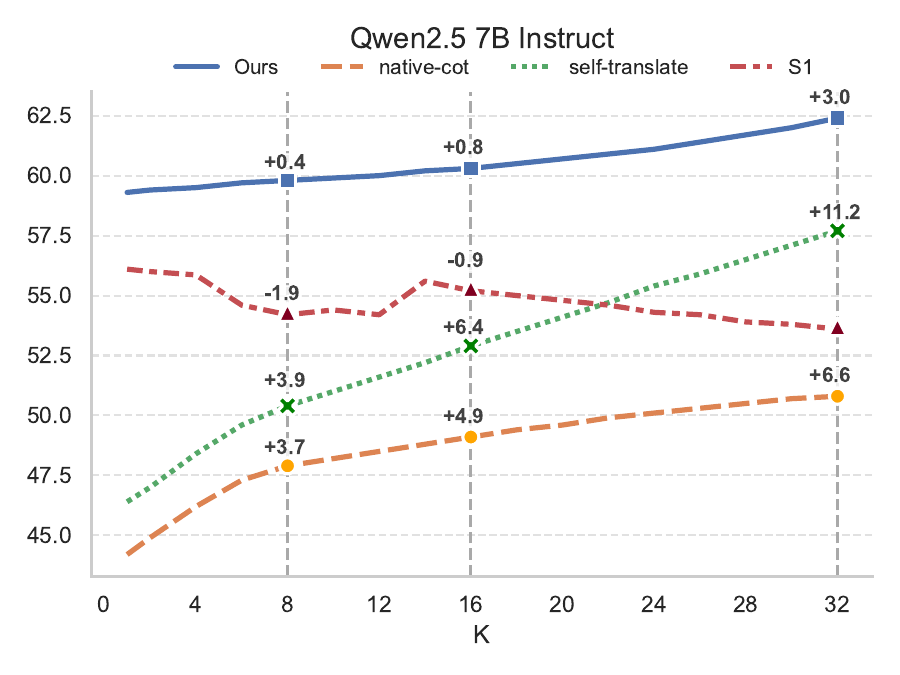}
\caption{\textbf{Average accuracy using Majority@k}\\performance of the Qwen-2.5 7B IT model on the MGSM benchmark by taking the Maj@k for $k \in [1\dots 32]$. Vertical dashed-lines indicate the relative gain from Maj@1 for k=\{8,16,32\} for each method.}
\label{fig:maj@k-qwen7b}
\end{figure}

We also analyze the consistency of our method on the final accuracy performance using using majority voting (Maj@k) with varying sample sizes. As shown in Figure~\ref{fig:maj@k-qwen7b}, models trained with our approach reach near-peak performance at relatively low \textit{k}: Maj@8 achieves over 95\% of the accuracy achieved at Maj@32, suggesting our method is optimizing for inference compute. All model sizes exhibit the same behavior, as we show in Appendix~\ref{app:maj@k-figure}. We observe high inconsistency in the results of S1 trained models at higher values of K, mainly in the medium- and low-resource languages. We also note that prompt-based baselines — particularly self-translation— continue to benefit from increasing \textit{k}, making it a reliable basis for our method.

\subsection{Generalization}  

\begin{table}[th!]
\small
\centering
\begin{adjustbox}{max width=1.0\columnwidth}
\begin{tabular}{rllll}
\bottomrule
\rowcolor{gray!15} 
\textbf{Model Size}& \textbf{Method} & \textbf{mGPQA} &\textbf{MMATH} & \textbf{PolyMath} \\
\toprule
\multirow{4}{*}{\textbf{1.5B}}&\textbf{base}&22.8&46.9& 7.5 \\ 
&\textbf{LIDR}& 18.8 \sneg{4.0} & 45.0 \sneg{1.9} & 5.9 \sneg{1.6} \textsuperscript{\textdagger} \\
&\textbf{S1}&24.5 \splus{1.7} & \textbf{53.8 }\splus{6.9}& 8.0 \splus{0.5}\\
&\textbf{self-cons (ours)}&\textbf{24.9} \splus{2.1} &48.8 \splus{1.8 }& 8.1 \splus{0.6} \\
\midrule
\multirow{4}{*}{\textbf{7B}}&\textbf{base}&24.6&71.1& 18.0 \\
&\textbf{LIDR}& 23.5 \sneg{1.1} & 67.0 \sneg{3.9} & 16.2 \sneg{1.8} \textsuperscript{\textdagger} \\
&\textbf{S1}&25.7 \splus{2.9} &\textbf{74.7} \splus{3.6} & 20.2 \splus{2.2}\\
&\textbf{self-cons (ours)}&\textbf{28.9} \splus{4.3} &74.3 \splus{3.2}& 20.5 \splus{2.5} \\
\midrule
\multirow{4}{*}{\textbf{14B}}&\textbf{base}&28.2&72.2& 17.5 \\
&\textbf{LIDR}&29.4 \splus{0.8} & 70.5 \sneg{1.7}& 15.2 \sneg{2.3}\textsuperscript{\textdagger}\\
&\textbf{S1}&32.1 \splus{3.9} &  77.4 \splus{5.2} & 19.5 \splus{2.0}\\
&\textbf{self-cons (ours)}&\textbf{32.9} \splus{4.7}&\textbf{78.4} \splus{6.2}&20.9 \splus{3.4}\\

\bottomrule
\end{tabular}
\end{adjustbox}
\caption{\textbf{Generalization of Qwen2.5-Instruct models on Out of Distribution (OOD) benchmarks}. Results on mGPQA, MMATH, and PolyMath (average of its 4 subtasks). Note that our models were trained on questions from GSM8K only. \textsuperscript{\textdagger} indicates best run results }
\label{tab:gen-domain}
\end{table}

\paragraph{Out-of-distribution Generalization} We evaluate our models trained with the GSM8k sample on three Out-of-distribution benchmarks: mGPQA, MMATH, and PolyMath. As shown in Table~\ref{tab:gen-domain}, our method achieves the best generalization performance. On mGPQA, we acheive an average gain of 2.1\%, 4.3\%, and 4.7\% for the 1.5B, 7B, and 14B models, respectively, surpassing that of S1-trained models. On MMath, our method acheives up to \textbf{6.2\%} gain in the 14B model. Conversely, LIDR has an adversarial impact on the performance, causing performance degradation on all OOD benchmarks. The PolyMath benchmark is the furthest from the GSM8k distribution, yet our method achieves the best relative performance gains. This corroborates the strength of our method in cross-domain generalization. We also observe that generalization benefits from model scale, indicating that larger models are more capable of leveraging our training signal. 
\label{sec:gen}

\paragraph{Cross-lingual Generalization} We train the \qwenone IT and \qwenseven models on a random subset of 1, 2, and 3 languages and evaluate their performances on the remaining, unseen languages.  As shown in Table~\ref{tab:gen2langs}, our method achieves substantial improvements in the unseen languages, \textbf{reaching up to 18.2\% for the \qwenseven-IT model}. Notably, when the model is trained on a single source language—e.g., Bengali—the performance gain in that language is maximized. Nonetheless, including additional languages during training enhances the model’s overall generalization to other unseen languages.
\begin{table}[t!]
\centering
\small
\begin{tabular}{crcc}
\bottomrule
\rowcolor{gray!15} 
\textbf{Model Size} & \textbf{L$_{src}$} & 
$\Delta_{seen}$ & $\Delta_{unseen}$\\ 
\toprule
\multirow{3}{*}{\textbf{1.5B}}&bn&16.8&4.0\\
&bn,fr&13.3&4.2\\
&de,es,th&13.8&6.2\\
\midrule
\multirow{3}{*}{\textbf{7B}}
&bn&23.8&15.6\\
&bn,fr&15.7&18.2\\
&de,es,th&19.6&15.2\\
\bottomrule
\end{tabular}
\caption{\textbf{Generalization to Unseen Languages on MGSM.} Models trained on $L_{\text{src}}$ languages are evaluated on held-out languages. $\Delta$: difference in performance from the base model.  
}
\label{tab:gen2langs}

\end{table}

\subsection{Different Target Distributions}

\label{sec:impact-dist}
We now analyze the strategies to estimate the target output distribution $q(y)$ used in reward modeling (see \S~\ref{sec:reward-modeling}). As shown in Table~\ref{tab:ablate-tgt-dist}, using the source language outputs to construct the target distribution yields the least improvement. However, it can be useful in cases where the model is initially indecisive, such as \qwenone. 

In contrast, constructing the target distribution from the outputs of English or combining the outputs of all languages achieves competitive performance. Specifically, using English outputs yields 1.4\% more improvements than the combined outputs for \qwenseven in MGSM. We hypothesize this is due to the less noisy English outputs and that English yields significantly higher base performance (at least 18\% above other languages). Nonetheless, we believe that the choice of combined output is favorable, especially in domains and models where English performance is not dominant or other languages perform competitively. In addition, the fact that our method performs well even when considering weaker languages makes it robust for wide applicability.

\subsection{Access to multilingual data for training}
\label{sec:translation-quality}

Our main approach accounts for domains where reasoning data is only available in one or a few languages by leveraging the translation capability of the model across supported languages. We now analyze the efficiency of our method in cases when multilingual data is accessible. 

We post-train our models on the Multilingual Grad School Mathematics (MGSM) dataset \cite{shi2022languagemodelsmultilingualchainofthought}. The dataset includes 250 examples from the test set of the GSM8k benchmark, manually translated into 10 languages besides English. We train on 25\% of the MGSM sampled uniformly over languages and evaluate on the remaining 75\%. As shown in Table~\ref{tab:train-mgsm-summarized}, this setup improves the results for the smaller models, but gets comparable average results for the largest model. We hypothesize that the largest model has enough quality to self-translate, and that, after all, it is the self-consistency of the trajectories that matters most, and that adequate translation of the questions suffices. 
\begin{table}[t!]
\centering
\small
\begin{adjustbox}{max width=1.0\columnwidth}
\begin{tabular}{rlc}
\bottomrule
\rowcolor{gray!15} 
\textbf{Model Size}& \textbf{Strategy} & \textbf{MGSM (avg)}\\
\toprule
\multirow{3}{*}{\textbf{1.5B}}&\textbf{Source Lang}&38.7 \splus{5.6}\\
&\textbf{English}&40.4 \splus{7.3}\\
&\textbf{Combined}&\textbf{40.9} \splus{7.8}\\
\midrule
\multirow{3}{*}{\textbf{7B}}&\textbf{Source Lang}&48.8 \splus{4.6} \\
&\textbf{English}&\textbf{61.0} \splus{16.8}\\
&\textbf{Combined}& 59.6 \splus{15.4}\\
\bottomrule
\end{tabular}
\end{adjustbox}
\caption{ \textbf{Performance of Qwen2.5-IT models trained with our method using different target distribution estimation strategies.} All models are trained on the monolingual data and evaluated on the MGSM benchmark.}
\label{tab:ablate-tgt-dist}
\end{table}

\section{Related Work}
\label{sec:related-work}

\paragraph{Multilingual Reasoning with Supervised Fine-Tuning.} Prior work addresses cross-lingual reasoning gaps by fine-tuning models on translated question pairs to improve alignment across languages \cite{zhu2024question,chen-etal-2024-breaking}. 
However, the scarcity of high-quality parallel data and the noise introduced by machine translation limit scalability and performance on complex reasoning tasks, making English-only reasoning (English-cot) remain the most effective baseline \citep{liu-etal-2025-translation, ranaldi-etal-2025-natural}.
Some works add a second step with preference optimization, which requires external translation, and use English trajectories as preferred with respect to native trajectories \cite{she-etal-2024-mapo,yang2025language} 

\paragraph{Multilingual Reasoning with Inference-Time Intervention.} To bypass data limitations, recent studies explore inference-time strategies adapted from monolingual reasoning. Longer chains of thought \citep{yong2025crosslingualreasoningtesttimescaling} and multilingual ensembling via majority voting \citep{gao2025thinkingmultilinguallyempowerllm} both improve cross-lingual performance, with English often serving as the most reliable reasoning language. As an alternative \cite{huang2024mindmergerefficientboostingllm,wang-etal-2025-bridging}, 
response generation can be manipulated using the logits to mimic those of English. Nevertheless, they perform poorly compared to supervised fine-tuning methods. 
\begin{table}[th!]
\centering
\small
\begin{adjustbox}{max width=1.0\columnwidth}
\begin{tabular}{rcc}
\bottomrule
\rowcolor{gray!15} 
\textbf{Model Size}& \textbf{Train Dataset} &\textbf{Acc. (avg)}\\
\toprule
\multirow{2}{*}{ \textbf{1.5B}}
& \textbf{GSM8k}&40.9 \splus{7.8}\\
& \textbf{MGSM}&43.5 \splus{10.4}\\
\midrule
\multirow{2}{*}{ \textbf{7B}}
& \textbf{GSM8k}& 59.3 \splus{15.1}\\
&\textbf{MGSM}&61.2 \splus{17.0}\\
\midrule
\multirow{2}{*}{ \textbf{14B}}
& \textbf{GSM8k}& 76.5 \splus{20.3}\\
&\textbf{MGSM}&76.5 \splus{20.3} \\
\bottomrule
\end{tabular}
\end{adjustbox}
\caption{\textbf{Impact of using manual translations in training Qwen2.5-IT models.} Results when training on a sample of the Multilingual data (MGSM, manual translations) compared to training on (GSM8K, self-translation before training). } 
\label{tab:train-mgsm-summarized}
\end{table}

\paragraph{Monolingual Reasoning.} Outcome-based reward modeling has recently shown that enforcing output consistency can improve reasoning without ground-truth supervision, but existing work is confined to English \citep{agarwal2025unreasonableeffectivenessentropyminimization, prabhudesai2025maximizing, he2026farunsupervisedrlvrscale}. We extend this paradigm to multilingual reasoning. 

\section{Conclusion and Future Work}

In this paper, we show that incentivizing cross-lingual consistency improves reasoning in the source language. Our method uses only monolingual questions, applying reinforcement learning with an unsupervised cross-lingual consistency loss to further train an existing model. It assumes no preferred language and enforces consistency across all languages. Experiments on MGSM show substantial gains, especially for weaker languages, while also improving stronger ones and reducing performance gaps. Results on OOD benchmarks indicate good generalization, though with smaller gains. Since our method relies only on monolingual questions, we hypothesize that training on more diverse domains could further enhance generalization.

In the future we would like to explore further domains, and extend our method to non-verifiable tasks. All things considered, our findings pave the way to improve the multilingual reasoning performance at large without any extra data annotation. 

\section*{Limitations}

Our experiments are limited to sub 20B LLMs, primarily due to compute constraints. While we observe that our method is increasingly effective as the size grows it would be interesting to explore its behaviour for larger models. In addition, we used the Qwen2.5 Models; It would be efficient to see the impact on more recent models such as Qwen3 \cite{yang2025qwen3technicalreport}.

Our training and evaluation were highly constrained by the data. The limited availability of reliable multilingual reasoning benchmarks—particularly for low-resource languages—restricts broader validation and remains a hindering challenge. We hope to address some of these limitations in future work.

\section*{Acknowledgments}

\bibliography{custom}

\appendix

\section{Self-consistency Training Curves}
    \begin{figure}[t!]
\includegraphics[width=1.\columnwidth]{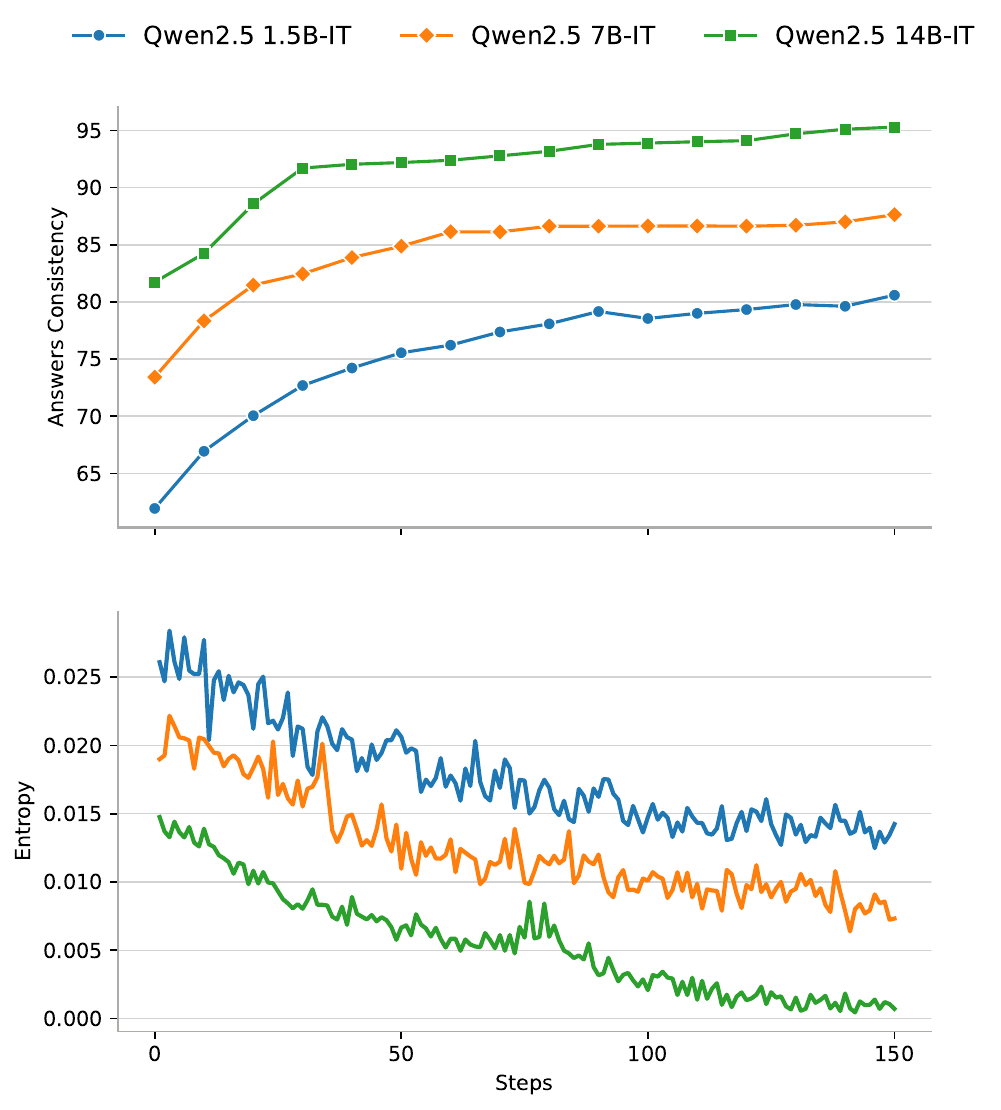}
\caption{\textbf{Training curves.} 
\textbf{Top:} Cross-lingual consistency improvement throughout training. \textbf{Bottom:} Entropy decreases throughout training.
}
\label{fig:perf-and-entropy-vs-steps}
\end{figure} 

\label{app:training-curves}
In Figure~\ref{fig:perf-and-entropy-vs-steps} (top), we show that our method yields a steady progression for all models. The progression improves steadily for all models, and the upward trend suggests further improvements could be achieved by scaling up the training compute. Concurrent to the improvements in cross-lingual consistency, we show in Figure~\ref{fig:perf-and-entropy-vs-steps} (bottom) that the entropy of the model's policy is steadily regularized as training progresses. This supports our view of the method as a cross-entropy regularization technique and links the improvement in cross-lingual consistency with the improvement of the model entropy.

\section{Self-consistency Accuracy Performances}

\subsection{Majority@k Performances}
\label{app:maj@k-figure}
    \begin{figure*}[t!]
\includegraphics[width=1.0\textwidth]{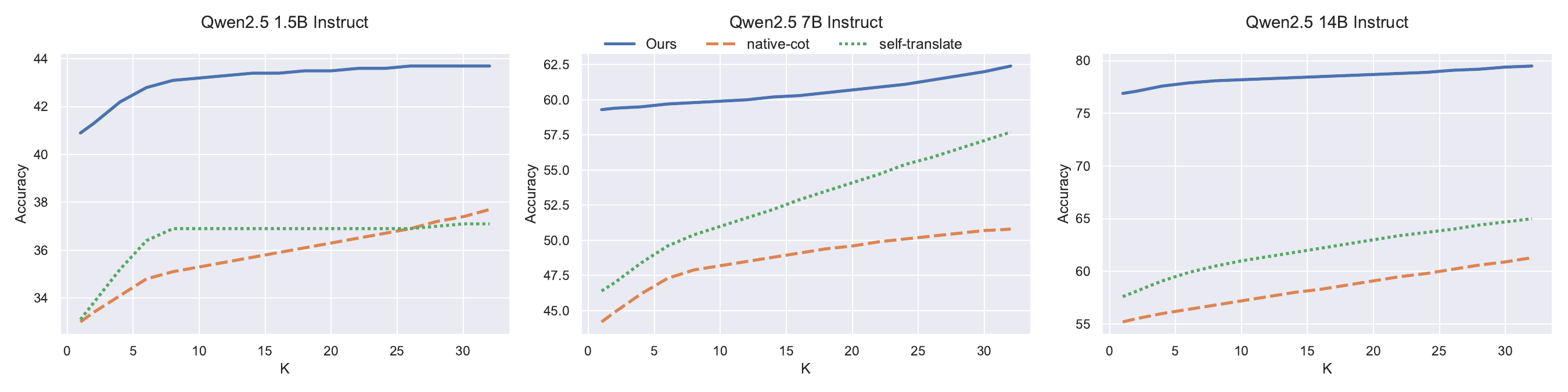}
\caption{\textbf{Accuracy Majority@k} performance of the Qwen-2.5 IT models on the MGSM benchmark by taking the Maj@k for $k \in [1 \dots 32]$}
\label{fig:maj@k-all-models}
\end{figure*} 

Figure~\ref{fig:maj@k-all-models} shows the performance of Qwen-2.5 IT models using Majority@k, for $k \in \{1,8,32\}$. Our method achieves near-peak performance at lower samples, making it more efficient with respect to inference budget.

 \subsection{Detailed Performances}
\begin{table*}[th!]
\small
\centering
\begin{adjustbox}{max width=1.0\textwidth}
\begin{tabular}{llcrrrrrrrrrrr!{\vrule}l}
\bottomrule
\rowcolor{gray!15} 
\textbf{Model} &\textbf{Method}&\textbf{K} & \textbf{bn} & \textbf{de} & \textbf{en} & \textbf{es} & \textbf{fr} & \textbf{ja} & \textbf{ru} & \textbf{sw} & \textbf{te} & \textbf{th} & \textbf{zh} & \textbf{avg}\\
\toprule

\multirow{12}{*}{\textbf{Qwen2.5-1.5B-Instruct}} & \multirow{3}{*}{\textbf{native-cot}} & 1 & 12.0 & 41.2 & 52.4 & 48.8 & 48.1 & 31.6 & 46.4 & 2.8 & 4.4 & 34.6 & 41.2 & 33.0 \\
&& 8 & 12.0 & 45.2 & 55.6 & 48.4 & 54.0 & 37.2 & 44.8 & 2.8 & 4.0 & 31.6 & 50.8 & 35.1 \\
&& 32 & 16.4 & 48.4 & 58.4 & 49.2 & 56.0 & 37 & 49.8 & 3.2 & 6.2 & 33.6 & 56.2 & 37.7 \\
\cmidrule(lr){2-15}
& \multirow{3}{*}{\textbf{self-translate}} & 1 & 14.4 & 44.2 & 52.4 & 50.0 & 42.1 & 23.6 & 38.8 & 8.4 & 12.4 & 36.8 & 40.8 & 33.1 \splus{0.1} \\
 && 8 & 14.2 & 48.0 & 56.8 & 49.0 & 54.8 & 36.8 & 48.4 & 3.2 & 6.0 & 32.8 & 56.0 & 36.9 \splus{1.8} \\
 && 32 & 8.0 & 45.2 & 
56.8 & 48.4 & 56.0 & 37.2 & 48.0 & 8.2 & 12.0 & 32.8 & 56.0 & 37.1 \splus{4.1} \\
\cmidrule(lr){2-15}
&\multirow{3}{*}{\textbf{eng-cot}} & 1 & 13.2 & 41.8 & 52.4 & 51.6 & 50.6 & 27.6 & 46.8 & 3.2 & 2.0 & 36.6 & 52.8 & 34.4 \splus{1.4} \\
&& 8 & 14.2 & 48 & 54.8 & 49.0 & 52.0 & 30.8 & 48.4 & 3.2 & 6.0 & 32.8 & 54.6 & 35.8 \splus{0.7} \\
&& 32 & 16.4 & 48.8 & 56.8 & 49.2 & 54.8 & 36.8 & 49.4 & 3.4 & 12.8 & 33.6 & 57.6 & 38.1 \splus{5.1} \\
\cmidrule(lr){2-15}
&\multirow{3}{*}{\textbf{ours}} & 1 & 21.2 & 53.6 & 50.3 & 63.2 & 56.4 & 36.0 & 55.2 & 5.2 & 7.2 & 48.0 & 54.0 & 40.9 \splus{7.9} \\
&& 8 & 24.0 & 56.4 & 56.0 & 66.0 & 57.2 & 39.0 & 57.6 & 4.8 & 7.2 & 48.0 & 58.4 & 43.1 \splus{8.0} \\
&& 32 & 24.4 & 56.8 & 56.4 & 67.2 & 57.2 & 39.8 & 60.2 & 5.2 & 7.2 & 48.4 & 58.4 & 43.7 \splus{10.7} \\
\midrule
\multirow{12}{*}{\textbf{Qwen2.5-7B-Instruct}} & \multirow{3}{*}{\textbf{native-cot}}& 1 & 30.0 & 60.4 & 78.6 & 40.2 & 48.2 & 49.8 & 37.4 & 19.8 & 18.2 & 54.8 & 49.2 & 44.2 \\
&& 8 & 30.8 & 62.4 & 80.6 & 57.2 & 51.6 & 50.4 & 44.2 & 21.6 & 20.0 & 53.4 & 54.6 & 47.9 \\
&& 32 & 31.2 & 63.2 & 88.4 & 58.6 & 55.2 & 52.8 & 48.0 & 23.6 & 21.2 & 56.2 & 60.8 & 50.8 \\
\cmidrule(lr){2-15}
&\multirow{3}{*}{\textbf{self-translate}}& 1 & 33.4 & 67.6 & 78.6 & 50.3 & 54.4 & 43.6 & 39.2 & 11.4 & 18.6 & 60.2 & 52.6 & 46.4 \splus{2.2} \\
&& 8 & 34.3 & 69.8 & 80.6 & 71.6 & 58.2 & 44.1 & 46.3 & 12.4 & 20.4 & 58.7 & 58.4 & 50.4 \splus{2.5} \\
&& 32 & 30.2 & 72.8 & 88.4 & 77.2 & 71.6 & 61.2 & 68.0 & 24.4 & 20.0 & 51.2 & 69.6 & 57.7 \splus{13.5} \\
\cmidrule(lr){2-15}
&\multirow{3}{*}{\textbf{eng-cot}} & 1 & 34.2 & 59.2 & 80.0 & 65.0 & 70.5 & 41.8 & 64.8 & 17.4 & 11.2 & 44.0 & 60.9 & 49.9 \splus{5.7} \\
&& 8 & 35.1 & 61.2 & 82 & 62.0 & 70.6 & 42.3 & 76.6 & 19.0 & 12.3 & 42.9 & 67.6 & 52.0 \splus{4.1} \\
&& 32 & 31.8 & 73.2 & 88.4 & 77.6 & 70.0 & 62.8 & 67.2 & 23.6 & 12.0 & 52.4 & 68.8 & 57.1 \splus{12.9} \\
\cmidrule(lr){2-15}
&\multirow{3}{*}{\textbf{ours}}& 1 & 54.8 & 75.2 & 81.8 & 62.8 & 71.6 & 55.6 & 76.0 & 17.6 & 24.0 & 73.2 & 59.6 & 59.3 \splus{15.1} \\
&& 8 & 56.3 & 77.7 & 80.4 & 62.0 & 70.2 & 56.3 & 76.2 & 19.2 & 22.0 & 71.3 & 66.1 & 59.8 \splus{11.9} \\
&& 32 & 58.2 & 78.0 & 82.6 & 57.0 & 74.6 & 60.8 & 78.8 & 24.6 & 24.8 & 78.4 & 68.6 & 62.4 \splus{18.2} \\
\midrule
\multirow{12}{*}{\textbf{Qwen2.5-14B-Instruct}} & \multirow{3}{*}{\textbf{native-cot}} & 1 & 37.6 & 61.6 & 84.2 & 75.2 & 48.8 & 47.2 & 44.0 & 40.4 & 21.8 & 66.0 & 80.8 & 55.2 \\
&& 8 & 38.6 & 63.6 & 86.3 & 76.4 & 52.2 & 47.8 & 46.0 & 44.1 & 24.0 & 64.3 & 82.0 & 56.8 \\
&& 32 & 40.2 & 64.6 & 94.8 & 80.2 & 56.8 & 52.0 & 56.4 & 48.2 & 28.4 & 68.4 & 84.0 & 61.3 \\
\cmidrule(lr){2-15}
& \multirow{3}{*}{\textbf{self-translate}} & 1 & 33.4 & 66.0 & 84.2 & 70.4 & 57.3 & 53.2 & 49.2 & 41.4 & 26.2 & 70.2 & 82.3 & 57.6 \splus{2.4} \\
&& 8 & 34.3 & 68.1 & 80.6 & 70.0 & 61.3 & 63.8 & 72.6 & 42.8 & 22.8 & 68.5 & 81.2 & 60.5 \splus{3.7} \\
&& 32 & 28.2 & 80.0 & 80.0 & 72.4 & 81.6 & 70.0 & 86.2 & 44.0 & 22.0 & 66.8 & 84.0 & 65.0 \splus{9.8} \\
\cmidrule(lr){2-15}
&\multirow{3}{*}{\textbf{eng-cot}} & 1 & 56.6 & 83.2 & 83.8 & 72.0 & 77.6 & 39.2 & 73.6 & 48.8 & 25.6 & 73.4 & 83.0 & 65.2 \splus{10.0} \\
&& 8 & 58.1 & 86.0 & 83.0 & 68.7 & 77.7 & 64.2 & 76.0 & 50.2 & 28.1 & 71.6 & 92.1 & 68.7 \splus{11.9} \\
&& 32 & 47.6 & 84.2 & 84.0 & 73.6 & 81.2 & 71.6 & 84.0 & 54.4 & 36.4 & 77.2 & 78.4 & 70.2 \splus{15.0} \\
\cmidrule(lr){2-15}
& \multirow{3}{*}{\textbf{ours}} & 1 & 65.6 & 89.2 & 88.4 & 87.8 & 82.2 & 80.8 & 88.4 & 49.6 & 44.6 & 81.6 & 88.0 & 76.9 \splus{21.7} \\
&& 8 & 67.4 & 92.2 & 86.9 & 86.7 & 80.6 & 81.8 & 90.6 & 54.2 & 48.0 & 82.6 & 88.4 & 78.1 \splus{21.3} \\
&& 32 & 66.2 & 90.0 & 89.2 & 90.4 & 84.0 & 82.6 & 91.6 & 54.2 & 49.8 & 84.6 & 92.4 & 79.5 \splus{24.3} \\

\bottomrule
\end{tabular}
\end{adjustbox}
\caption{\textbf{Accuracy for Majority@K for base and models trained with our method on the MGSM benchmark.} }
\label{tab:acc-at-k}
\end{table*}

\label{app:maj@k}
In Table~\ref{tab:acc-at-k}, we show the performance of different methods by computing the accuracy with majority vote at different sample sizes (Maj@k). Ideally, a self-consistent model achieves the best performance at a lower sample size, hence a more efficient inference token consumption. Models trained with our approach achieve their highest performance at lower K values, with Maj@8 achieving more than 95\% of that of Maj@32. 

In addition, results show that the performance of prompt-based methods, specifically self-translation, improves consistently with higher K, corroborating our assumption that they provide a reliable basis for our method.

\section{Detailed Performances}
Tables~\ref{tab:gen-domain} --~\ref{tab:train-mgsm} show the detailed performances of our method training.

\begin{table*}[th!]
\small
\centering
\begin{adjustbox}{max width=1.0\linewidth}
\begin{tabular}{llrrrrrrrrrrr!{\vrule}l}
\bottomrule
\rowcolor{gray!15} 
& \textbf{Method} & \textbf{bn} & \textbf{de} & \textbf{en} & \textbf{es} & \textbf{fr} & \textbf{ja} & \textbf{ru} & \textbf{sw} & \textbf{te} & \textbf{th} & \textbf{zh} & \textbf{avg}\\
\toprule
\multirow{3}{*}{\textbf{\qwenone IT}}
&\textbf{base}&17.0&20.1&\textbf{28.5}&21.1&23.1&20.1&\textbf{27.2}&2.2&5.9&20.3&25.7&22.8\\
&\textbf{english-cot}&\textbf{22.3}&\textbf{25.0}&28.5&23.7&21.0&\textbf{23.0}&18.5&\textbf{12.9}&\textbf{13.7}&21.7&22.1&21.2 \sneg{1.6} \\
&\textbf{self-cons (ours)}&21.8&22.7&26.9&\textbf{26.3}&\textbf{25.7}&21.7&26.8&2.3&6.2&\textbf{24.1}&\textbf{27.6}& \textbf{24.9} \splus{2.1}\\
\midrule
\multirow{3}{*}{\textbf{\qwenseven IT}}
&\textbf{base}&22.8&25.7&30.4&27.7&33.3&24.3&26.3&8.3&16.5&23.4&32.1&24.6\\
&\textbf{english-cot}&26.1&\textbf{31.9}&30.4&32.1&27.9&\textbf{32.6}&\textbf{29.0}&\textbf{16.2}&\textbf{27.2}&26.1&32.4& 28.4 \splus{3.8}\\
&\textbf{self-cons (ours)}&\textbf{26.7}&31.0&\textbf{32.4}&\textbf{32.5}&\textbf{39.9}&\textbf{27.7}&28.8&12.2&19.6&\textbf{29.2}&\textbf{37.7}&\textbf{28.9 } \splus{4.3} \\
\midrule
\multirow{3}{*}{\textbf{\qwenfrtn IT}}
&\textbf{base}&23.4&35.7&32.8&31.9&32.1&27.2&33.3&15.0&17.6&27.5&34.2&28.2\\
&\textbf{english-cot}&\textbf{34.8}&37.5&32.8&\textbf{37.9}&36.7&29.7&32.0&19.9&19.6&\textbf{32.9}&\textbf{40.4}& 32.2 \splus{4.0}\\
&\textbf{self-cons (ours)}&29.5&\textbf{39.1}&\textbf{36.6}&33.5&\textbf{37.3}&\textbf{32.1}&\textbf{34.6}&\textbf{23.9}&\textbf{25.7}&31.5&38.4& \textbf{32.9} \splus{4.7}\\

\bottomrule
\end{tabular}
\end{adjustbox}
\caption{\textbf{Generalization to other domains}. Results on mGPQA for our models (which were trained on questions from GSM8K).}
\label{tab:mgpqa-detailed}
\end{table*}

\begin{table*}[th!]
\small
\centering
\begin{adjustbox}{max width=1.0\linewidth}
\begin{tabular}{llrrrrrrrrrrr!{\vrule}l}
\bottomrule
\rowcolor{gray!15} 
& \textbf{Strategy} & \textbf{bn} & \textbf{de} & \textbf{en} & \textbf{es} & \textbf{fr} & \textbf{ja} & \textbf{ru} & \textbf{sw} & \textbf{te} & \textbf{th} & \textbf{zh} & \textbf{avg}\\
\toprule
\multirow{3}{*}{\textbf{\qwenone IT}}
&\textbf{Source Lang}&30.8&48.4&55.2&52.4&53.6&35.6&52.0&4.0&4.8&40.8&48.8& 38.7 \splus{5.6}\\
&\textbf{English}&\textbf{38.8}&47.6&\textbf{57.8}&53.6&\textbf{59.6}&32.0&50.4&4.4&7.2&44.0&48.8& 40.4 \splus{7.3}\\
&\textbf{Combined}&21.2&\textbf{53.6}&50.3&\textbf{63.2}&56.4&\textbf{36.0}&\textbf{55.2}&\textbf{5.2}&\textbf{7.2}&\textbf{48.0}&\textbf{54.0}& \textbf{40.9} \splus{7.8}\\
\midrule
\multirow{3}{*}{\textbf{\qwenseven IT}}
&\textbf{Source Lang}&43.2&68.8&66.0&52.0&58.4&50.2&48.4&17.2&22.6&57.6&52.8&48.8 \splus{4.6} \\
&\textbf{English}&54.0&\textbf{79.6}&80.8&62.6&\textbf{74.0}&\textbf{60.0}&69.6&\textbf{20.0}&\textbf{34.4}&\textbf{75.2}&\textbf{60.8}& \textbf{61.0} \splus{16.8}\\
&\textbf{Combined}&\textbf{54.8}&75.2&\textbf{81.8}&\textbf{62.8}&71.6&55.6&\textbf{76.0}&17.6&24.0&73.2&59.6& 59.6 \splus{15.4}\\
\bottomrule
\end{tabular}
\end{adjustbox}
\caption{ \textbf{Performance of different target distribution estimation strategies.} All models are trained on the monolingual data and evaluated on the MGSM benchmark. 
}
\label{tab:ablate-tgt-dist}
\end{table*}

\begin{table*}[th!]
\small
\centering
\begin{adjustbox}{max width=1.0\linewidth}
\begin{tabular}{llrrrrrrrrrrr!{\vrule}l}
\bottomrule
\rowcolor{gray!15} 
& \textbf{Train Dataset} & \textbf{bn} & \textbf{de} & \textbf{en} & \textbf{es} & \textbf{fr} & \textbf{ja} & \textbf{ru} & \textbf{sw} & \textbf{te} & \textbf{th} & \textbf{zh} & \textbf{avg}\\
\toprule
\multirow{2}{*}{ \textbf{\qwenone IT}}
& \textbf{GSM8k}& 21.2& 53.6&50.3& 63.2& 56.4& 36.0& 55.2& 5.2& 7.2& 48.0& 54.0&40.9 \splus{7.8}\\
& \textbf{MGSM}&29.2&53.2&56.8&63.0&56.8&41.6&55.2&5.6&6.4&50.0&61.2&43.5 \splus{10.4}\\
\midrule
\multirow{2}{*}{ \textbf{\qwenseven IT}}
& \textbf{GSM8k}&54.8&75.2&81.8&62.8&71.6&55.6&76.0&17.6&24.0&73.2&59.6& 59.3 \splus{15.1}\\
&\textbf{MGSM}&63.2&78.8&78.5&72.0&64.4&62.4&68.0&21.2&32.8&75.6&62.8& 61.2 \splus{17.0}\\
\midrule
\multirow{2}{*}{ \textbf{\qwenfrtn IT}}
& \textbf{GSM8k}&65.6&89.2&88.4&87.8&82.2&80.8&88.4&49.6&44.6&81.6&88.0& 76.9 \splus{21.7}\\	
&\textbf{MGSM}&62.0&86.4&88.0&90.4&85.2&82.8&88.2&47.6&41.6&83.6&85.2& 76.5 \splus{20.3}\\
\bottomrule
\end{tabular}
\end{adjustbox}
\caption{\textbf{Impact of using manual translations in training.} Results when training on a sample of the Multilingual data (MGSM, manual translations) compared to training on (GSM8K, self-translation before training). Both settings are evaluated on the questions of the holdout sample of the MGSM dataset (75\%). } 
\label{tab:train-mgsm}
\end{table*}

\section{Comparison to S1 trained models}

Models trained with S1 test-time scaling in English exhibit impressive improvements in other languages \citep{yong2025crosslingualreasoningtesttimescaling}. They achieve competetive performances to our method. We analyze the reasoning traces of the S1 models and ours to better compare the methods.

\subsection{Reasoning Traces}
\label{app:quote-and-think}
\begin{example}[Comparing the reasoning trace of the s1's \textit{quote-and-think} vs ours]
{box:quote-and-think}

\textbf{Question}: Los patos de Janet ponen 16 huevos por día. Ella come tres en el desayuno todas las mañanas y usa cuatro para hornear magdalenas para sus amigos todos los días. Vende lo que sobra en el mercado de productores diariamente a \$2 el huevo fresco de pato. ¿Cuánto gana en dólares todos los días en el mercado de productores?

(Janet's ducks lay 16 eggs per day. She eats three for breakfast every morning and uses four to bake muffins for her friends every day. She sells the leftover eggs at the farmers market daily for \$2 per fresh duck egg. How much does she earn in dollars each day at the farmers market?)
\bigskip

\textbf{s1's trace}: \\
Okay, let me try to figure this out. So, Janet has some ducks that lay eggs. Let's see, they lay 16 eggs per day. Then, Janet uses some of them herself and bakes with some, and sells the rest at the market. The question is how much she earns each day from selling the leftover eggs. Each egg sells for \$2 ... Wait, that seems straightforward ... so I think \$18 is the correct answer. Respuesta: Janet's patos ponen 16 huevos diariamente. Ella consume 3 huevos para el desayuno y utiliza 4 huevos para hornear magdalenas, totalizando 7 huevos usados diariamente. Por lo tanto, los huevos sobrantes para vender son 16 - 7 = 9 huevos. Vendiendo cada huevo a \$2, Janet gana 9 x \$2 = \$18 diariamente en el mercado de productores. \textbf{Respuesta Final \textbackslash boxed\{18\}}

\medskip
\textbf{Self-consistency (ours) trace}: \\
Primero, Janet tiene 16 huevos por día. Come 3 huevos y usa 4 huevos, por lo que en total usa 3 + 4 = 7 huevos. Esto deja 16 - 7 = 9 huevos para vender. Si vende cada huevo a \$2, entonces gana 9 * 2 = \$18 todos los días en el mercado de productores.\textbf{Respuesta: \$18}
\label{quote-n-think}
\end{example}

In Box 1, we show traces of Qwen2.5 14B trained with our method and the s1-method on a Spanish question. Models trained with S1 test-time scaling technique exhibit a quote-and-think paradigm where they do reasoning first in English and deliver the final response --and occasionally short chain of thought-- in the prompt language. Albeit efficient, this does not imply that the S1 training recipe induces better multilingual reasoning, but rather surfaces the model's translation capabilities as a proxy for enhanced reasoning.

\subsection{Test-time Token Consumption}

\begin{table*}[th!]
\small
\centering
\begin{adjustbox}{max width=1.0\linewidth}
\begin{tabular}{lrrrrrrrrrrr!{\vrule}l}
\bottomrule
\rowcolor{gray!15} 
& \textbf{bn} & \textbf{de} & \textbf{en} & \textbf{es} & \textbf{fr} & \textbf{ja} & \textbf{ru} & \textbf{sw} & \textbf{te} & \textbf{th} & \textbf{zh} & \textbf{avg}\\
\toprule
\textbf{\qwenfrtn IT}&229&189&127&139&146&157&122&142&314&212&197&188.3\\
\quad \textbf{Ours}& 241&188&135&155&246&151&173&152&251&198&167&216.1\\	
\quad \textbf{S1}&430&358&270&440&296&356&320&340&714&410&290&366.1\\
\bottomrule
\end{tabular}
\end{adjustbox}
\caption{\textbf{Average response token length on the MGSM dataset.} Our models surpass the performance of S1 at much lower token length.} 
\label{tab:resp-token-len}
\end{table*}

\label{app:token-budget}
S1 models tend to do longer reasoning in English, while our method arrives at the same answer with shorter and language-consistent chain-of-thought. In Table ~\ref{tab:resp-token-len}, we show the average and maximum token length of the responses per language using \qwenfrtn trained with our method and with S1 test-time scaling.


\end{document}